\documentclass[runningheads]{llncs}
\usepackage[T1]{fontenc}
\usepackage{graphicx}
\usepackage{hyperref}
\usepackage{color}

\usepackage{booktabs}
\usepackage{listings}

\begin{document}
\title{Small but Significant: On the Promise of Small Language Models for Accessible AIED}
\titlerunning{Small but Significant}
%
\author{Yumou Wei\orcidID{0009-0002-1364-8300} \and
Paulo Carvalho\orcidID{0000-0002-0449-3733} \and
John Stamper\orcidID{0000-0002-2291-1468}}
%

\authorrunning{Y. Wei et al.}
%
\institute{Human-Computer Interaction Institute, Carnegie Mellon University, USA \\ \email{\{yumouw,pcarvalh,jstamper\}@andrew.cmu.edu}}
%
\maketitle              
\begin{abstract}
GPT has become nearly synonymous with large language models (LLMs), an increasingly popular term in AIED proceedings. A simple keyword-based search reveals that 61\% of the 76 long and short papers presented at AIED 2024 describe novel solutions using LLMs to address some of the long-standing challenges in education, and 43\% specifically mention GPT. Although LLMs pioneered by GPT create exciting opportunities to strengthen the impact of AI on education, we argue that the field's predominant focus on GPT and other resource-intensive LLMs (with more than 10B parameters) risks neglecting the potential impact that small language models (SLMs) can make in providing resource-constrained institutions with equitable and affordable access to high-quality AI tools. Supported by positive results on knowledge component (KC) discovery, a critical challenge in AIED, we demonstrate that SLMs such as Phi-2 can produce an effective solution without elaborate prompting strategies. Hence, we call for more attention to developing SLM-based AIED approaches.

\keywords{Small Language Models  \and Accessible AIED \and Knowledge Component Discovery.}
\end{abstract}

\section{Introduction}
It is an exciting time for AIED. Technological breakthroughs in large language models (LLMs)~\cite{brown2020language} have provided unprecedented opportunities for AIED researchers and practitioners to solve some of the long-standing challenges in the field~\cite{KASNECI2023ChatGPT}. The excitement is aptly exemplified by the community's fast adoption of LLMs in AIED research---of the 76 long and short papers accepted for AIED 2024, 61\% (47 papers) describe innovative solutions using LLMs, as revealed by a simple keyword-based search in the proceedings~\cite{olney_artificial_2024}. Among the ever-expanding constellation of available LLMs, the GPT family, including ChatGPT~\cite{openai2022chatgpt} and GPT-4~\cite{openai2023gpt4}, appears to be the community's favorite: 33 of the 47 papers (70\%) adopting LLMs also mention GPT. Although LLMs pioneered by GPT herald exciting possibilities to reinforce AI's positive influence on education, we argue that the community's predominant focus on GPT and other similar resource-intensive gigantic language models (with more than ten billion parameters) risks neglecting the critical impact that \textbf{small language models} (SLMs) can make in creating equitable and accessible education central to the mission of AIED. 

The definition of SLMs is constantly changing as new technologies emerge to shape the landscape of language models. The BERT model~\cite{devlin2019bert} in its largest configuration, for example, has 340 million parameters---an overwhelming amount in 2018 but only a fraction by today's standard. In relation to the current state of the art, we consider a language model small if it has fewer than ten billion parameters and requires modest hardware resources, such as a consumer-grade GPU. Canonical examples of SLMs include Llama-2 7B~\cite{touvron2023llama2}, Mistral 7B~\cite{jiang2023mistral}, and Phi-2~\cite{hughes_phi-2_2023}. Phi-2, a lightweight but capable model that has only 2.7B parameters, might be a particularly good fit for the AIED community and the range of problems we are trying to address. Trained on high-quality ``textbook-like'' data~\cite{gunasekar2023textbooks}, Phi-2 subsumes deep knowledge about various academic disciplines and aligns better with educational contexts, which require precision and reliability, than other SLMs trained on mixed-quality data sourced from the Internet. Its smaller size also enables local deployment on consumer-grade hardware, desirable for most educational settings where computational resources are limited.

Educational institutions operate under distinct constraints that make their AI implementation needs different from those of commercial environments. Budget limitations, technical infrastructure, privacy requirements, and equity considerations all influence technology adoption in educational settings~\cite{Reich2017}. GPT-scale LLMs typically require substantial computational resources for local deployment or incessant API costs for cloud access, not affordable to all teachers or students~\cite{KASNECI2023ChatGPT}.
SLMs, however, only require a fraction of the resources entailed by LLMs and can be deployed on modest hardware at a much lower cost---Phi-2's 2.7 billion parameters only require about 5.4 GB of memory for storage with a 16-bit representation of floating-point numbers\footnote{2.7B float numbers require $16\times$2.7B=43.2B bits, which translate to 43.2B/8B = 5.4 gigabytes if 8 bits make 1 byte.}, which can fit comfortably to a consumer-grade GPU. 

One argument that justifies the higher costs of GPT-scale LLMs is their superior performance in various tasks. However, we argue that the more affordable and accessible SLMs can also deliver impressive results if we manage to exploit their potential adequately. In Section~\ref{sec: kc}, we present a case study of knowledge component (KC) discovery~\cite{kli}, a critical challenge in AIED, and describe our unique solution using Phi-2.
Our approach makes creative use of Phi-2 as a probability machine to measure question similarity and applies a clustering algorithm to identify questions belonging to the same KC; results on two datasets show that instructors can better predict student performance using the KCs generated by our approach than using those produced by experts or the more powerful GPT-4o. 
These positive findings from the case study reinforce our position that \textbf{small language models such as Phi-2 can provide effective solutions to critical AIED problems and hold great promise as a catalyst for inclusive, personal, and ethical education equitably accessible to teachers and students}, as AIED 2025 advocates. 

\section{Background}

\subsection{The Rise of Large Language Models in Education}

The field of education has tremendously benefited from the advances in natural language processing (NLP) in recent decades, which have evolved from rule-based approaches to statistical methods and eventually to neural-network models~\cite{Litman_2016}. Early educational applications used relatively simple NLP techniques for tasks such as automated essay scoring~\cite{Shermis2013}; more recent work, however, uses advanced language models to tackle increasingly complex challenges in education.

Introduced in 2017, the Transformer architecture~\cite{vaswani2017attention} enables researchers to build more sophisticated language models with enhanced language understanding and generation capabilities. Together with more efficient hardware and better available corpora, this architectural innovation spurred the development of models with progressively larger parameter counts---some prominent milestones include GPT-3~\cite{brown2020language} (175B parameters), PaLM~\cite{chowdhery2022palm} (540B parameters), and GPT-4~\cite{openai2023gpt4} (estimated 1.76T parameters). These gigantic language models have demonstrated remarkable capabilities across various educational applications, including but not limited to hint creation~\cite{pardos2023learning}, question generation~\cite{Sarsa2022}, and KC discovery~\cite{Moore2024}. 

Concomitant to the development of more capable models is the emphasize of \emph{scaling}---increasing model size, training data, and computational resources---as the primary mechanism for improving model performance~\cite{kaplan2020scalinglawsneurallanguage}. This scaling law suggests that many unexpected capabilities can emerge as model size increases, with larger models generally outperforming smaller ones across diverse tasks~\cite{wei2022emergent}. While the successful application of the scaling law has nearly depleted the available benchmarks to measure the progress of LLMs, urging the development of the ``Humanity's Last Exam''\footnote{\url{https://agi.safe.ai/}}, it has also raised the computational and financial requirements that prevent resource-constrained educational institutions from equitably using LLMs, and necessitated stricter, more private access to the source code and training data that could have helped researchers build more effective AIED tools. Moreover, the community's widespread predilection for large and even larger models can exacerbate the danger of overlooking the impact that SLMs can make in providing effective and accessible AIED solutions. 

\subsection{The Potential of Small Language Models in Education}

In contrast to the scaling efforts, researchers have also developed smaller and more efficient models that challenge the dominance of scaling as the only way to attain good performance. More recently, models like Phi-2 (2.7B parameters) have demonstrated that careful data curation and innovative training methodologies can produce surprisingly capable models at significantly smaller scales~\cite{hughes_phi-2_2023}.

Developed by Microsoft Research, Phi-2 is an epitome of efficient language models. This SLM is built on the standard Transformer decoder-only architecture and is trained with the conventional next-token prediction objective. What makes it special, however, is not architectural innovations but the unique training methodology used. Unlike many larger models trained on vast but heterogeneous corpora sourced from the Internet, Phi-2 was trained predominantly on what the researchers call ``textbook-quality data''~\cite{gunasekar2023textbooks}---carefully curated content with an emphasis on educational materials, synthetic texts designed for reasoning capabilities, and filtered web content with high educational value. 

This unique training methodology, which ranks data quality higher than quantity, results in an efficient SLM that is particularly useful for educational applications. In competitive benchmarks that evaluate reasoning skills in math (GSM8k~\cite{cobbe2021gsm8k}) and coding (HumanEval~\cite{chen2021evaluating}, MBPP~\cite{austin2021program}), Phi-2 substantially outperformed Mistral 7B~\cite{jiang2023mistral} and Llama-2 13B~\cite{touvron2023llama2}, which are $1.6\times$ and $3.8\times$ larger than Phi-2. Compared to the $25\times$ larger Llama-2 70B~\cite{touvron2023llama2}, Phi-2 achieved significantly better performance in coding and demonstrated comparable reasoning skills in math~\cite{hughes_phi-2_2023}. In the MMLU benchmark~\cite{hendrycks2021measuring}, which assesses language model knowledge in 57 academic subjects, Phi-2 outperformed Llama-2 13B (54.8) and achieved a score (56.7) comparable to that achieved by Mistral 7B (60.1).

From a computational efficiency perspective, Phi-2 also offers distinct advantages for educational applications. Requiring approximately 5.4 GB of memory for storage (with additional memory for inference), Phi-2 can be deployed on consumer-grade hardware with modest requirements (the conventional 16-GB GPU), enabling local inference without cloud infrastructure dependencies. This flexibility in deployment helps reduce the first digital divide~\cite{Attewell2001} that prevents resource-constrained schools from using the latest AI tools, and protects student privacy~\cite{Prinsloo2017} by not requiring student data to be shared with a third party. 

Phi-2's solid results on academic benchmarks and modest requirements on computer hardware make it a competitive alternative to gigantic language models that entail substantial computational resources and provoke critical privacy concerns. Its extensive pre-training on high-quality textbook-like data makes Phi-2 particularly tuned to educational applications. For example, as shown in Table~\ref{tab: phi-2}, when we ask Phi-2 to generate the most probable tokens one after another with \emph{an effectively empty prompt}, Phi-2 by design refers to an educational Wikipedia page (with a valid link\footnote{\url{https://en.wikipedia.org/wiki/List\_of\_minor_planets:\_386001\%E2\%80\%93387000}}) and spontaneously \emph{generates questions} related to an entity mentioned on the page. In what follows, we describe a concrete case study in which we creatively used Phi-2 to design a KC discovery algorithm that outperformed instructional experts and its GPT counterpart. 

\begin{table}[t]
    \centering
    \caption{Phi-2 is particularly tuned to educational applications. Configured to select the most probable token (top-1 sampling) at each step, it starts to refer to educational resources and generate questions when asked to write after the \texttt{<|endoftext|>} token. } 
    \label{tab: phi-2}
    \begin{tabular}
{p{0.99\textwidth}}
    \vspace{-1em}
\begin{lstlisting}[language=Python, frame=lines,basicstyle=\ttfamily\small,showstringspaces=false]
<|endoftext|>

Paragraph:<url_begin>https://en.wikipedia.org/wiki/List
%20of%20minor%20planets%3A%20386001%E2%80%93387000<url_end>

Topic: Science

1. What is the Minor Planet Center?
Answer: The Minor Planet Center is an organization that 
collects and disseminates data on minor planets, comets, 
and other small Solar System bodies.
\end{lstlisting} \\ [-3em]
    \end{tabular}
\end{table}


\section{Case Study: Knowledge Component Discovery}\label{sec: kc}

Representing specific concepts or skills that students acquire through learning to perform a task or solve a problem, knowledge components (KCs) are essential elements in the KLI framework~\cite{kli} that help instructors assess student learning. Traditionally, instructional experts are elicited to participate in Cognitive Task Analysis (CTA)~\cite{Clark2008} to identify the KCs associated with each assessment item, but CTA incurs considerable time and labor cost even when applied to moderately sized question banks. The accelerating adoption of AI in education aggravates the burden on instructors, who are overwhelmed by the growing amount of AI-generated questions that each needs to be analyzed by hand.

To address this challenge, a recent approach~\cite{Moore2024} uses GPT-4~\cite{openai2023gpt4} to extract KCs from multiple-choice questions (MCQs). The authors devised elaborate prompting strategies to ask GPT-4 to simulate instructional experts or textbook authors. Although in an evaluation study, the majority of the three participants preferred GPT-generated KCs to those designed by experts for more than 60\% of the evaluated questions, this approach produced KC labels with slightly different wording for questions that instructors think should belong to the same KC~\cite{Moore2024}. In our replication of their study using more advanced GPT-4o, the most intelligent non-reasoning LLM offered by OpenAI, we obtained 614 unique KC labels for 630 MCQs from the same e-learning dataset\footnote{\url{https://pslcdatashop.web.cmu.edu/DatasetInfo?datasetId=5426}} used by the authors~\cite{Moore2024}. The large number of KC labels comparable to the number of questions suggests that some labels can be merged. In fact, we discovered that GPT-4o had produced unnecessarily refined labels (e.g., ``Analyze CTA'', ``Analyze CTA in E-learning'', and ``Analyze CTA methodologies'') that could have been merged. 

In our recent work~\cite{wei2025kclusterllmbasedclusteringapproach}, we demonstrate that exploiting the native potential of a language model as a ``probability machine'' rather than the more conventional text generation capabilities can lead to a strong KC discovery algorithm even with SLMs such as Phi-2. The core idea is that language models can induce a novel measure of question similarity, which a clustering algorithm can use to identify groups of similar questions that are likely to share the same KC. Inspired by word collocations, we postulate that if one question increases the likelihood of another question appearing, the two questions are congruent and likely to relate to the same KC. We derive a formula for question congruity, our novel measure of question similarity mathematically equivalent to the pointwise mutual information (PMI)~\cite{church-hanks-1989-word} between two questions, and describe an algorithm that uses Phi-2 to calculate various required probabilities. 

We evaluated our approach against instructional experts and our replication of the previous study~\cite{Moore2024} using more advanced GPT-4o, on two datasets collected in a graduate e-learning course taught by two different instructors in 2022 and 2023\footnote{\url{https://pslcdatashop.web.cmu.edu/DatasetInfo?datasetId=5843}}. A common practice to compare different KC discovery approaches is to fit an Additive Factors Model (AFM)~\cite{afm} with the KCs generated by each method to student response data; a better KC discovery approach should allow an instructor to predict student responses with a lower root mean square error (RMSE). On the 2022 dataset, our approach achieved an RMSE of $0.4220$, outperforming both experts ($0.4235$) and GPT-4o ($0.4395$); likewise, on the 2023 dataset, our approach scored $0.4066$, leading both experts ($0.4075$) and GPT-4o ($0.4101$). Notably, GPT-4o, a highly capable LLM, performed the worst on the two distinct datasets; this strengthens our claim that SLMs can also deliver superior results if their potential is adequately exploited. 


\section{Conclusion}
Through this forward-looking paper, we did not argue that the AIED community should eschew LLMs in favor of their more efficient counterparts, nor did we suggest that SLMs are capable of everything LLMs can do. Similar to many AIED researchers, we share the excitement about the complementary development of both lines of NLP research and their potential application to education. However, \textbf{to empower teachers and students for an equitable future as AIED 2025 advocates, the promise of SLMs in providing accessible AIED solutions is not to be neglected.} As shown in the case study, an innovative exploitation of SLM's potential can deliver superior results than the standard use of LLMs based on intensive prompt engineering. We urge the AIED community to reconsider, next time when making a convenient API call to an LLM, whether it endangers the accessibility to the target audience, who may actually benefit from an SLM.

\bibliographystyle{splncs04}
\bibliography{mybibliography}
%





\end{document}